\documentclass{article}
\pdfoutput=1

\usepackage[preprint,nonatbib]{neurips_2024}

\usepackage[utf8]{inputenc} %
\usepackage[T1]{fontenc}    %
\usepackage{hyperref}       %
\usepackage{url}            %
\usepackage{booktabs}       %
\usepackage{amsfonts}       %
\usepackage{nicefrac}       %
\usepackage{microtype}      %
\usepackage{xcolor}         %

\usepackage{graphicx}

\usepackage{amsmath}
\usepackage{cuted}
\usepackage{wrapfig}

\usepackage{tabularx}
\usepackage{subcaption}
\usepackage{tabulary,multirow,overpic,subfloat}

\newcommand{\app}{\raise.17ex\hbox{$\scriptstyle\sim$}}

\newcolumntype{x}[1]{>{\centering\arraybackslash}p{#1pt}}
\newcolumntype{y}[1]{>{\raggedright\arraybackslash}p{#1pt}}

\newlength\savewidth
\newcommand{\tablestyle}[2]{\setlength{\tabcolsep}{#1}\renewcommand{\arraystretch}{#2}\centering\footnotesize}
\makeatletter\renewcommand\paragraph{\@startsection{paragraph}{4}{\z@}
  {.5em \@plus1ex \@minus.2ex}{-.5em}{\normalfont\normalsize\bfseries}}\makeatother

\usepackage{placeins} %

\definecolor{myblue}{HTML}{1a3375}
\definecolor{mygreen}{HTML}{74b90c}
\definecolor{myred}{HTML}{962c37}
\definecolor{myyellow}{HTML}{FFBC42}
\definecolor{mypurple}{HTML}{9055A2}

\DeclareRobustCommand\onedot{\futurelet\@let@token\@onedot}
\def\@onedot{\ifx\@let@token.\else.\null\fi\xspace}
\def\eg{\emph{e.g}\onedot}

\makeatother

\usepackage{orcidlink}

\begin{document}

\title{\textbf{\textcolor{myblue}{X}-\textcolor{mygreen}{V}\textcolor{myred}{I}\textcolor{myyellow}{L}\textcolor{mypurple}{A}}: Cross-Modality Alignment for\\Large Language Model}

\author{Hanrong Ye\textsuperscript{1,2}\thanks{Work done during an internship at NVIDIA.},
De-An Huang\textsuperscript{1},
Yao Lu\textsuperscript{1},
Zhiding Yu\textsuperscript{1},
Wei Ping\textsuperscript{1},
Andrew Tao\textsuperscript{1}, \\
\textbf{Jan Kautz\textsuperscript{1},
Song Han\textsuperscript{1,3},
Dan Xu\textsuperscript{2},
Pavlo Molchanov\textsuperscript{1},
Hongxu Yin\textsuperscript{1}}
\\
NVIDIA\textsuperscript{1} \quad HKUST\textsuperscript{2} \quad MIT\textsuperscript{3}
}

\maketitle

\begin{figure}[h]
    \centering
    \vspace{-25pt}
    \includegraphics[width=\textwidth]{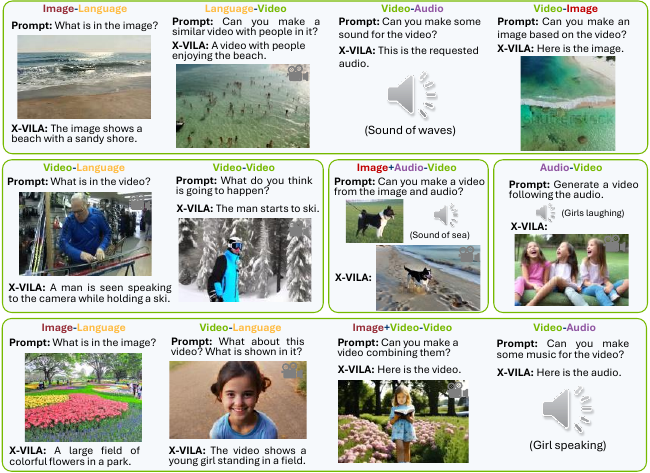}
    \vspace{-15pt}
    \caption{
    We introduce \textbf{\textcolor{myblue}{X}-\textcolor{mygreen}{V}\textcolor{myred}{I}\textcolor{myyellow}{L}\textcolor{mypurple}{A}}, a foundation model for \textbf{\textcolor{myblue}{cross-modality}} understanding, reasoning, and generation in the domains of \textcolor{mygreen}{\textbf{video}}, \textcolor{myred}{\textbf{image}}, \textcolor{myyellow}{\textbf{language}}, and \textcolor{mypurple}{\textbf{audio}}.
    X-VILA demonstrates the ability to perceive~(see, hear, and read) multi-modality inputs, and generate~(draw, speak, and write) multi-modality responses. Conversations are continuous within each green box. Best viewed in color.
    \label{fig:teaser}}
\vspace{-2mm}
\end{figure}

\begin{abstract}
  We introduce X-VILA, an omni-modality model designed to extend the capabilities of large language models (LLMs) by incorporating image, video, and audio modalities.
  By aligning modality-specific encoders with LLM inputs and diffusion decoders with LLM outputs, X-VILA achieves cross-modality understanding, reasoning, and generation. 
  To facilitate this cross-modality alignment, we curate an effective interleaved any-to-any modality instruction-following dataset.
  Furthermore, we identify a significant problem with the current cross-modality alignment method, which results in visual information loss.
  To address the issue, we propose a visual alignment mechanism with a visual embedding highway module. 
  We then introduce a resource-efficient recipe for training X-VILA, that exhibits proficiency in any-to-any modality conversation, surpassing previous approaches by large margins. X-VILA also showcases emergent properties across modalities even in the absence of similar training data.
  The project will be made open-source.
\end{abstract}

\section{Introduction}
Large language models (LLMs) provide an emerging foundation for enhancing various deep learning tasks beyond the realm of natural language processing. As an example, research community has been quickly extending the fast progress of LLMs~\cite{bert,raffel2020exploring,dai2019transformer,OpenAI_ChatGPT,touvron2023llama, touvron2023llama2,alpaca,vicuna2023,karamcheti2021mistral,penedo2023refinedweb,chowdhery2022palm,yi,qwen} towards the computer vision (CV) 
 domain~\cite{liu2023llava,alayrac2022flamingo,driess2023palm,chen2023pali,li2023blip,fuyu,bai2023qwen,GPT4,zhu2023minigpt}. 
The introduction of LLMs in CV tasks enables vision models to perform many zero/few-shot and in-context learning tasks that are ``promptable'' through user questions, potentially empowering reasoning capabilities for the first time.
Despite remarkable progress, cross-modality alignment is still a challenging task as the joint training stage for cross-modality learning requires carefully designed feedback signal~\cite{wei2021finetuned,Dai2023InstructBLIP} to guide the connected foundation models~\cite{alayrac2022flamingo,liu2023llava,li2023blip}, backed by cross-modality datasets at scale~\cite{zhu2023multimodal,kakaobrain2022coyo-700m,schuhmann2022laion}. 
Hence, the majority of existing studies revolve around a solitary input modality linked to LLMs, with the output being solely text. For example, Flamingo~\cite{alayrac2022flamingo}, LLaVA~\cite{liu2023llava}, and VILA~\cite{lin2023vila} delve into image input, while Video-LLaMA~\cite{zhang2023videollama} and LITA~\cite{huang2024lita} specifically concentrates on video input.
Exploring the integration of various modalities into a cohesive framework is a crucial yet relatively unexplored research challenge~\cite{tang2024codi,wu2023next,lu2022unified} in the domain of multi-modality LLMs, yet observed practical in proprietary GPT-4o~\cite{GPT4}.

This study focuses on the development of a systematic approach to integrate multiple modalities, such as video, image, and audio, into an LLM at both the input and output stages. Our objective is to facilitate cross-modal conversations in an any-to-any modality (or ``X-to-X'') LLM, allowing for generation in different modalities. 
To accomplish the ambitious objective, we present a two-phase alignment mechanism: 
\textbf{\textit{(i) Textual alignment.}}
We align input and output representations of different modalities to the textual embedding space of the LLM~\cite{wu2023next}.
Specifically, in regard to the input of LLM, we use a unified embedding space that allows for the sharing of features extracted from encoders across diverse modalities. As for the output of LLM, we employ fine-tunable modality-specific diffusion models to convert the generated outputs of the LLM into content that aligns with the respective modalities.
\textbf{\textit{(ii) Visual alignment.}}
We observe that the previous textual alignment alone fails to preserve visual features adequately in vision-to-vision generation tasks, such as image-to-video and video-to-image generation. This limitation can be attributed to the loss of information during the projection process from visual encoders to the LLM, as well as the LLM's tendency to prioritize common concepts over specific visual details. 
To address this issue, we propose a new module named Visual Embedding Highway~(VEH). The VEH module facilitates the direct guidance of visual decoders by enabling visual features to bypass the LLM. By incorporating VEH, we have observed a notable enhancement in the correspondence of visual content between the input and output stages of our framework.

On the other hand, the scarcity of cross-modality instruction-following data poses a significant challenge in the development of any-to-any modality (or ``X-to-X'') LLMs. This limitation severely restricts the progress in creating LLMs that can seamlessly handle multiple modalities in both input and output ends. 
Existing datasets provide limited data, mostly in the form of X-to-text or text-to-X. 
Therefore, we curate a new X-to-X dataset based on multi-modality data from WebVid~\cite{Bain21webvid} and ActivityNet Captions~\cite{krishna2017denseANC} to facilitate cross-modality interactions between text, audio, image, and video. Overall, we synthesize more than 1.5M multi-modality conversations, with each conversation containing at least one cross-modality question-and-answer pair.

To achieve the cross-modality input-output alignment of LLMs in our X-to-X LLM, we design three major training phases: 
\textbf{(i)} A data-effective alignment phase that involves aligning the multi-modality encoders with the LLM inputs and the multi-modality decoders with the LLM outputs. 
\textbf{(ii)} An interleaved multi-modality pre-training phase with interleaved instruction data across modalities for enhanced in-context learning performance. 
\textbf{(iii)} An X-to-X cross-modality instruction tuning phase that includes a two-step alignment process: textual alignment and visual alignment mechanism.
Through our innovative approach to multi-modality alignment, we build a powerful X-to-X multi-modality LLM with the ability to comprehend and generate multi-modality content. We term our new model ``X-VILA'' for \textbf{cross-modality} understanding, reasoning, and generation in the domains of \textbf{V}ideo, \textbf{I}mage, \textbf{L}anguage, and \textbf{A}udio.
For instance, as shown in Figure~\ref{fig:teaser} and Figure~\ref{fig:seattle}, X-VILA demonstrates its capacity to recognize the subjects in the image, which results from our vision-language alignment training.
Then, it can retrieve its knowledge and make logical deductions to answer the user's questions about the content in the image. Last but not least, it can generate aligned multi-modality output that matches the given context.

In summary, this work makes contributions in three aspects:
\begin{itemize}
    \item A new family of any-to-any modality chat LLM that is able to conduct multi-modality conversations by understanding signals from different modalities and generating content in various formats, including video, audio, image, and text.
    \item A novel 2-step alignment mechanism that effectively aligns both semantic and visual details between the input and output spaces. This mechanism ensures a comprehensive and accurate correspondence between the input and output of our X-to-X LLM. 
    \item The creation of a new X-to-X multi-modality instruction tuning dataset that is proven effective for cross-modality alignment. This dataset serves as a valuable resource for future research in the realm of multi-modality foundation models.
\end{itemize}

\section{Methodology}

\begin{figure}[t]
    \centering
    % \vspace{-20pt}
    \includegraphics[width=\textwidth]{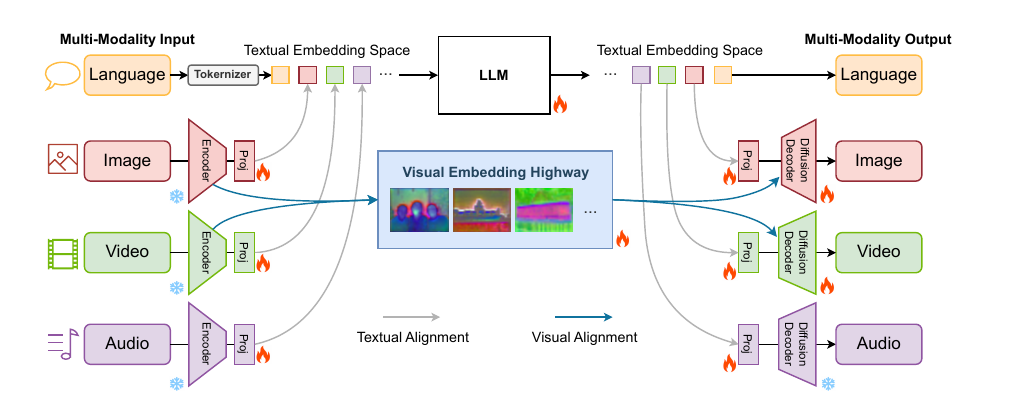}
    \caption{X-VILA schematic diagram. 
    X-VILA augments a pretrained LLM towards new modalities via (i) connecting pretrained encoders to LLM input textual embedding space and (ii) connecting pretrained diffusion decoders to the LLM output textual embedding space (Section~\ref{sec:x-vila-arch}). The system is jointly trained via a new cross-modality alignment procedure (Section~\ref{sec:x-vila-training}).}
    \label{fig:schematic}
% \vspace{-15pt}
\end{figure}

\subsection{X-VILA Architecture}
\label{sec:x-vila-arch}
We consider four common modalities in this work: text, image, video, and audio.
The tenet of X-VILA is an alignment-based architecture to augment an LLM with the ability to ``see/hear/read'' multi-modality inputs and ``draw/speak/write'' multi-modality outputs, as shown in Figure~\ref{fig:schematic}.

\noindent \textbf{Modality-specific encoders.}
We adopt modality-specific encoders to handle inputs from different modalities. This strategy harvests the pre-trained understanding ability of the domain expert encoders and has been proven successful in many vision-language models~\cite{alayrac2022flamingo,li2023blip,liu2023llava}. To better align embeddings of different modalities, we use ImageBind encoders~\cite{Girdhar2023ImageBindOE}, which unify features from different modalities, including image, video, and audio, into one feature space.
Technically, for each modality $m \in \{\text{`text', `image', `video', `audio'} \}$, we notate the encoders as $\text{\textbf{Enc}}_{m}$. 
For text modality, the encoder is a text tokenizer~\cite{kudo2018sentencepiece}, while for other modalities they are ImageBind transformers~\cite{Girdhar2023ImageBindOE}.
We then use modality-specific trainable linear layers, notated as $\textbf{P}^{\text{in}}_m$, to project $\text{\textbf{Enc}}_{m}$ output into embedding sequences $\mathbf{S}$ in the textual embedding space of the following LLM. We can formulate this process as:
\begin{equation}
    \mathbf{S}^{\text{in}} = \{ \textbf{P}^{\text{in}}_m (\textbf{Enc}_{m} (\mathbf{X}_m)) \},
\end{equation}
where $\mathbf{X}_m$ is input from different modalities $m\in \{\text{`text', `image', `video', `audio'} \}$.

\noindent \textbf{Large language model (LLM).} 
LLM serves as the ``brain'' of our framework. It processes information from the textual embedding space and predicts language outputs correspondingly. We adopt Vicuna-7B-1.5~\cite{vicuna2023,touvron2023llama2}, which demonstrates state-of-the-art language understanding and generation ability. For easier understanding, we slightly abuse the annotation and write the autoregressive process of generating output embedding sequence $\mathbf{S}^{\text{out}}$ by the LLM as:
\begin{equation}
    \mathbf{S}^{\text{out}} = \mathbf{LLM} ( \mathbf{S}^{\text{in}}).
\end{equation}

\noindent \textbf{Modality-specific decoders.}
For generating multi-modality outputs other than text, we adopt the ``modality-specific generation tokens'' designed by~\cite{wu2023next}. 
Other than common text tokens, there are three types of modality-specific generation tokens: image generation tokens \{[IMG$_i$], $i\in [1,N_{img}]\}$, video generation tokens \{[VID$_i$], $i\in [1,N_{vid}]\}$, and audio generation tokens \{[AUD$_i$], $i\in [1,N_{aud}]\}$. $N_{img}$, $N_{vid}$, and $N_{aud}$ are the numbers of generation tokens for image, video, and audio, respectively.
These modality-specific generation tokens are added to the vocabulary of LLM. The LLM is trained to predict when to generate these modality-specific generation tokens, and these generation tokens will be translated for the synthesis of image, video, or audio, via a set of modality-specific decoders~(\textit{i.e.}, generation models). 
Technically, we extract the subset of output embedding sequence $\mathbf{S}^{\text{out}}$ corresponding to the aforementioned generation tokens of modality $m$. We name this subset the generation embedding sequence $\mathbf{S}^{\text{gen}}_m$.
We use modality-specific transformer layers, denoted as output projection layers $\textbf{P}^{\text{out}}_m$, to project $\mathbf{S}^{\text{gen}}_m$ to the feature space of the original pre-trained text encoder of the modality-specific decoder. As the resulting embedding will be used to control the modality-specific decoder via cross-attention, we name the resulting embedding vector as ``textual controller embedding'' $\mathbf{E}^{\text{text}}_m$. Thus we have:
\begin{equation}
\mathbf{E}^{\text{text}}_m = \textbf{P}^{\text{out}}_m (\mathbf{S}^{\text{gen}}_m).
\end{equation}
\cite{wu2023next} freezes the decoder models and only supervises the $\mathbf{E}^{\text{text}}_m$ to be similar to the original text encoders of the diffusion models. This behavior largely limits the synergy between generation models and the other parts of the model, as the learning target is essentially to mimic the pre-trained text encoder of the diffusion models. 
Differently, we include the modality-specific decoder models in fine-tuning to better align them with the LLM and other parts of the unified generative framework. The training details will be discussed in a later section.
Specifically, to achieve a better multi-modality generation ability, we employ state-of-the-art generation models trained on large-scale data as modality-specific decoders. We adopt VideoCrafter2 \cite{chen2024videocrafter2} for video generation, Stable Diffusion 1.5~\cite{rombach2022ldm} for image generation, and AudioLDM~\cite{liu2023audioldm} for audio generation.

\begin{wrapfigure}{r}{0.5\textwidth}
\vspace{-15pt}
  \begin{center}
    \includegraphics[width=.5\textwidth]{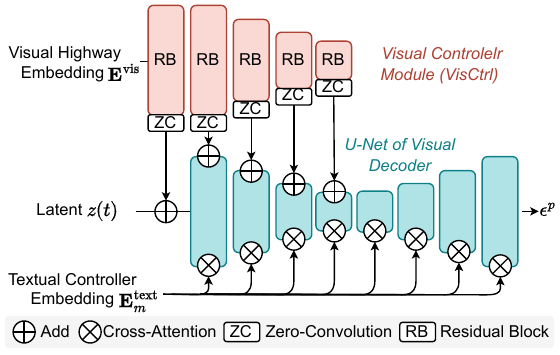}
  \end{center}
  \vspace{-0.3cm}
   \caption{Illustration of the proposed visual embedding highway in X-VILA. The visual highway embedding $\mathbf{E}^{\text{vis}}$ is obtained from the visual encoder. The design incorporates a visual controller module responsible for processing $\mathbf{E}^{\text{vis}}$ and generating control signals. These signals are then incorporated into various layers of the U-Net in visual decoders.
   $\mathbf{E}^{\text{text}}_m$ is ``textual controller embedding'', which is the subset of output embedding sequence $\mathbf{S}^{\text{out}}$ corresponding to the generation tokens of modality $m$.
   $z(t)$ is the latent at reverse step $t$.
   $\epsilon^p$ is the predicted noise by U-Net.
   }
  \label{fig:vhn}
  \vspace{-0.3cm}
\end{wrapfigure}

\begin{wrapfigure}{r}{0.5\textwidth}
\vspace{-0.3cm}
  \begin{center}
     \includegraphics[width=0.5\textwidth]{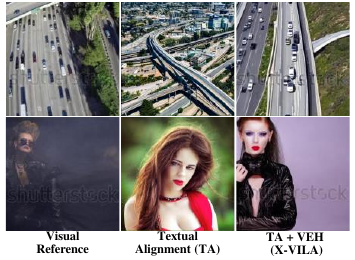}
     \vspace{-18pt}
  \end{center}
   \caption{Effectiveness of the proposed visual embedding highway network. Given the visual reference image/video, we prompt the model with ``\texttt{Please generate an image similar to the semantics in the input.}'' Compared to textual alignment only (TA), our visual embedding highway (VEH) helps preserve visual details from the visual inputs.}
  \label{fig:veh_motivation}
  \vspace{-0.2cm}
\end{wrapfigure}

\noindent \textbf{Visual embedding highway.}
The weakness of the previously introduced text-space-based alignment is the inadequate visual features available at the output end, as can be seen in examples in Figure~\ref{fig:veh_motivation}. 
Intuitively, this stems from the one-to-many correspondence between text and visual semantic spaces, \eg, ``a dress'' may relate to images varying in colors and styles.

To address this issue, we propose a visual embedding highway that bridges the visual encoders and decoders, built to alleviate the information loss when projecting high-dimensional visual content to the textual embedding space.
Specifically, we obtain the layer-wise feature maps from the ImageBind visual encoder and add up these features as visual highway embedding  $\mathbf{E}^{\text{vis}}$. $\mathbf{E}^{\text{vis}}$ has shape $H\times W\times C$, where $H$ and $W$ are height and width of the feature maps, $C$ is the embedding vector.
To control the decoder using $\mathbf{E}^{\text{vis}}$, we design a light-weight visual controller~(VisCtrl) module based on the philosophy of~\cite{mou2023t2i,zhang2023adding} to process $\mathbf{E}^{\text{vis}}$.
The controller module comprises 4 stages, where each stage consists of two residual convolutional blocks. These blocks have cascading spatial dimensions that match the resolution settings in the U-Net encoder~\cite{rombach2022ldm} of image/video decoders. 
In each stage, there is an additional convolutional block initialized with zero weights. 
This block generates output control signals for the stage, which are initially zero at the start of the training.
These control signals are added to different stages of the U-Net, as shown in Figure~\ref{fig:vhn}. Inspired by~\cite{xiao2023fastcomposer}, we employ a conditioning rate $\alpha \in [0,1]$ to regulate the proportion of steps conditioned by visual features.
Therefore, the noise prediction process in each reverse step $t$ in the visual decoders can be written as:
\begin{equation}
\epsilon^p =
\begin{cases} 
\text{U-Net}_m (z(t), \text{VisCtrl}_m(\mathbf{E}^{\text{vis}}), \mathbf{E}^{\text{text}}_m) & \text{if } t < T \times \alpha  \\
\text{U-Net}_m (z(t), \text{Null}, \mathbf{E}^{\text{text}}_m) & \text{if } t \geq T \times \alpha  
\end{cases}, m \in \text{\{`image', `video'\}}.
\label{eq:epsilon}
\end{equation}
where $\epsilon^p$ is the predicted noise given input latent $z(t)$, $T$ is the number of diffusion steps, $\text{U-Net}_m$ is the U-Net of the diffusion decoder for modality $m$, and $\text{VisCtrl}_m$ is the visual control module for modality $m$.``Null'' means no VEH feature is passed to the U-Net at the corresponding timestep.
During instruction tuning process on X-to-X datasets, both the U-Net and the controller modules are fine-tuned together. This manner ensures a better synergy between decoders and the LLM.

The experimental results introduced in the later sections show that the proposed visual embedding highway can significantly increase the consistency between the generation results and the visual context of our multi-modality unified generation model.

\subsection{X-VILA Training}
The training process of X-VILA is divided into three phases, namely (i) encoder-LLM-Decoder alignment training, (ii) interleaved data pre-training, and (iii) X-to-X cross-modality instruction fine-tuning. 
We describe the details of X-VILA training in Appendix~\ref{sec:x-vila-training} due to space limit.

\section{Experiments}
\label{sec:results}
\subsection{Datasets and Evaluation}
\noindent\textbf{Setup.} In this work, we utilize different datasets for different training phases. For the first encoder-LLM-decoder alignment training, the X-text pairs from LLaVA-pretrain~\cite{liu2023llava}, cc3m~\cite{sharma2018conceptual}, WebVid~\cite{Bain21webvid}, AudioCaps~\cite{audiocaps}, and WavCaps~\cite{mei2023wavcaps} are utilized.
During the interleaved data pre-training phase, we construct interleaved multi-modality corpus from  MMC4~\cite{zhu2023multimodal} and ActivityNet Captions~\cite{krishna2017denseANC}.

In the final X-to-X cross-modality instruction tuning, we create \textbf{a new X-to-X dataset} from WebVid~\cite{Bain21webvid} and ActivityNet Captions. We synthesize conversation samples in 6 types based on the modalities in input and output ends: video-to-image, video-to-video, image-to-video, video-to-audio, audio-to-video, image\&audio-to-video.
Statistically, for ActivityNet Captions, we make 10009 image-to-video, 10009 video-to-image, and 10009 video-to-video conversations.
For WebVid, we randomly select 500k training samples and build 499,915 image-to-video, 499,915 video-to-image, 499,915 video-to-video, 32,874 audio-to-video, 32,874 video-to-audio, and 32874 image+audio-to-video conversations. Each conversation contains more than one pair of cross-modality Q\&A pairs. Some conversation examples are shown in Figure~\ref{fig:x2xdataset}.
We blend our X-to-X dataset with SFT datasets from LLaVA~\cite{liu2023llava}, VideoChat~\cite{li2023videochat}, NextGPT-instructions~\cite{wu2023next}, and Alpaca~\cite{alpaca}.

\begin{figure}[t]
    \centering
    % \vspace{-10pt}
    \includegraphics[width=\textwidth]
    {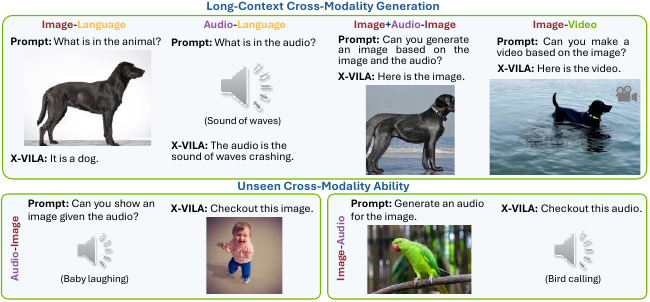}
    \caption{We observe emergent abilities of X-VILA without training on similar data: \textbf{(i) Long-context cross-modality generation ability.} Combine multiple inputs from different modalities and generate consistent content.
    \textbf{(ii) New types of cross-modality ability.} Conduct image-to-audio and audio-to-image generation tasks. Conversations are continuous left-to-right within each green box. 
    }
    \label{fig:emergent}
\end{figure}

\noindent\textbf{Evaluation.} For benchmarking the X-to-X alignment ability of different models, we randomly sample a validation subset of WebVid and ActivityNet Captions to build the cross-modality conversations for evaluation. Specifically, for ActivityNet Captions we generate 100 video-to-image, 100 video-to-video, and 100 image-to-video conversations.
For WebVid, we curate 100 video-to-image, 100 image-to-video, 100 video-to-video, 62 audio-to-video, 62 image+Audio-to-video, and 62 audio-to-video  conversations for evaluation.
In order to evaluate the similarity between ground-truth annotations and model predictions across different modalities, we introduce a metric called the ``X-to-X Alignment Score~(X$^2$A Score)''. To compute this score, we employ the ImageBind transformer~\cite{Girdhar2023ImageBindOE} to extract embedding vectors from the audio, video, and image predictions as well as the corresponding ground truths. We then calculate the cosine similarity scores between these vectors. The resulting scores are presented as percentages, ranging from 0 to 100. Finally, we average the scores across all validation samples to obtain the X$^2$A scores for each type of data. 

\noindent\textbf{Baseline methods.}
We conduct a comparison between our model and Next-GPT~\cite{wu2023next}, a recently introduced instruction-following LLM designed for multi-modality understanding and generation. Their method is restricted to textual alignment exclusively.

\begin{table}[t]
% \vspace{-10pt}
  \centering
  \resizebox{.7\linewidth}{!}{
  \tablestyle{4pt}{1.2}\begin{tabular}{l | ccc }
  \toprule
  \multicolumn{1}{c|}{\textbf{Method}} &  \textbf{VID2IMG}  & \textbf{VID2VID}  & \textbf{IMG2VID}    \\
  \midrule
  Next-GPT~\cite{wu2023next} & 27.85	&  10.47	&  13.08\\
  \textbf{X-VILA} w/ X2X text & 36.09	 & 46.18 & 	45.93 \\
  \textbf{X-VILA} w/ X2X text + VEH (img) & \textbf{44.06} & 	46.68 & 45.94 \\
  \textbf{X-VILA} w/ X2X text + VEH (img+vid) & 43.95 & \textbf{49.76} & \textbf{48.81} \\
  \bottomrule
  \end{tabular}
  }
  \vspace{2pt}
   \caption{X$^2$A scores on ActivityNet Captions. ``w/ X2X text'' denotes using our X-to-X dataset for textual alignment only. ``VEH (img)'' denotes using the proposed visual embedding highway (VEH) for image decoder, while ``VEH (img+vid)'' denotes using VEH for both image and video decoders. We observe that image generation task is significantly improved after using VEH (img), and the video generation tasks are boosted after using  VEH on video decoder.}
% \vspace{-25pt}
\label{tab:x2x_an}
\end{table}

\begin{table}[t]
  \huge
  % \vspace{-20pt}
  \centering
  \resizebox{1.\linewidth}{!}{
  \tablestyle{3pt}{1.2}\begin{tabular}{l | ccc ccc }
    \toprule
  \multicolumn{1}{c|}{\textbf{Method}}  &  \textbf{VID2IMG} & \textbf{IMGAUD2VID} & \textbf{VID2AUD} & \textbf{IMG2VID} & \textbf{VID2VID} & \textbf{AUD2VID}    \\
  \midrule
  Next-GPT~\cite{wu2023next} & 15.31 & 44.63 & 8.17 & 	38.23 & 31.81 & 37.13 \\
  \textbf{X-VILA} w/ X2X text  & 53.82 & 49.54 & 22.79 & 42.94 & 44.42 & 42.23 \\
  \textbf{X-VILA} w/ X2X text + VEH (img)  & 67.40 & 	48.64	 & 23.53 & 42.66 & 	43.04 & 42.04 \\
 \textbf{X-VILA} w/ X2X text + VEH (img+vid)  & \textbf{67.94} & 	\textbf{59.71} & \textbf{23.87}	 & \textbf{57.01} & \textbf{57.39} &  \textbf{49.44} \\
  \bottomrule
  \end{tabular}}
  \vspace{2pt}
   \caption{X$^2$A scores on WebVid. ``w/ X2X text'' denotes using our X-to-X dataset for textual alignment only. ``VEH (img)'' denotes using the proposed visual embedding highway (VEH) for image decoder, while ``VEH (img+vid)'' denotes using VEH for both image and video decoders. The effectiveness of visual embedding highway is solid for image and video generation.}
\vspace{-10pt}
\label{tab:x2x_webvid}
\end{table}

\begin{figure}[t]
    \centering
    % \vspace{-15pt}
    \includegraphics[width=1\textwidth]{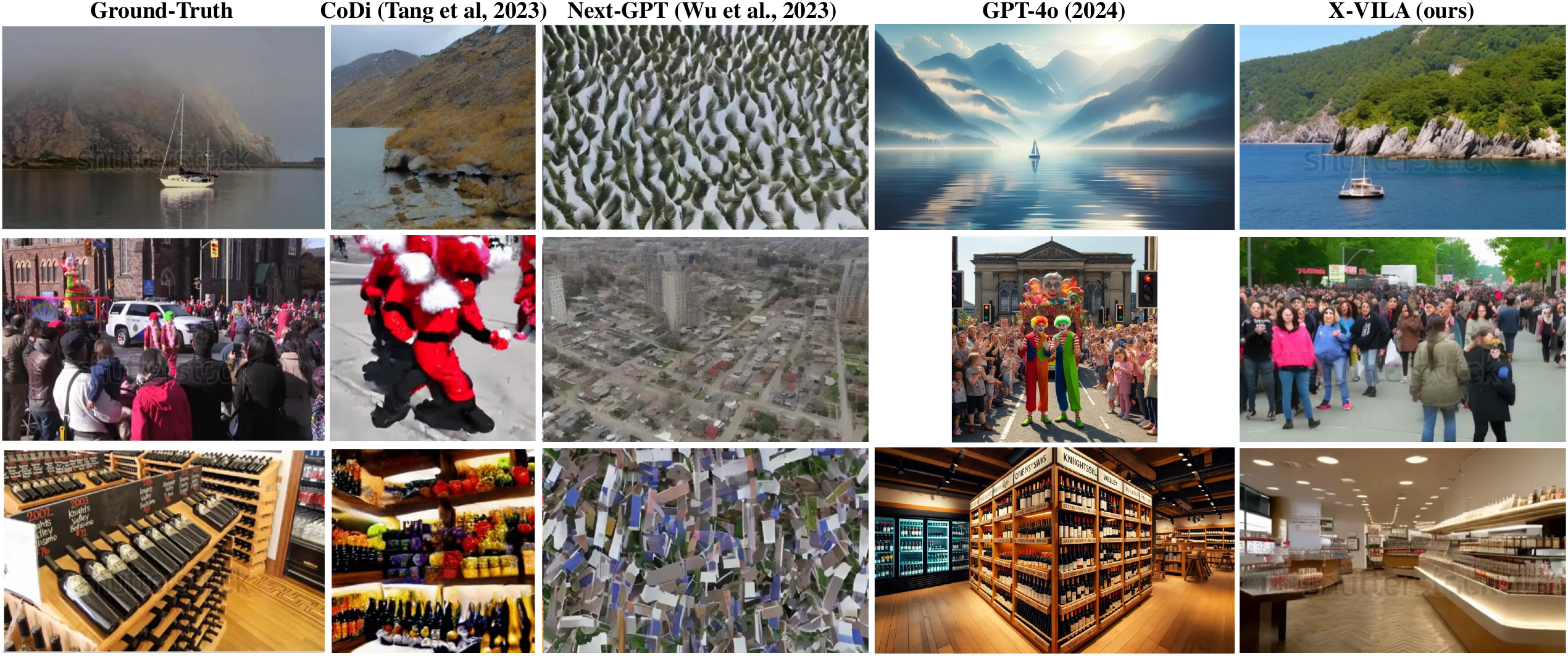}
    \caption{Visual comparison to the recent any-to-any modality LLMs including Next-GPT~\cite{wu2023next}, CoDi~\cite{tang2024codi}, and GPT-4o~\cite{openai2024chatgpt4} on the cross-modality alignment task to generate a video similar to the input image context. X-VILA demonstrates good generation quality and better visual cross-modality consistency. GPT-4o is only able to generate images but not videos. 
   }
    \label{fig:sota_compare}
% \vspace{-10pt}
\end{figure}

\begin{figure}[t]
    \centering
    % \vspace{-20pt}
    \begin{subfigure}[b]{.25\textwidth}
    \includegraphics[width=\textwidth]{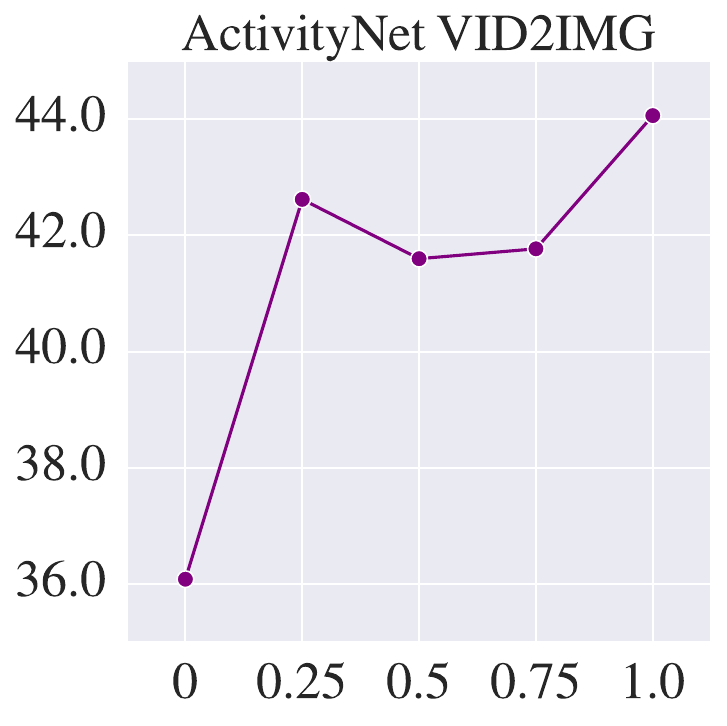}
    \vspace{7pt}
    \end{subfigure}
    \begin{subfigure}[b]{.25\textwidth}
    \includegraphics[width=\textwidth]{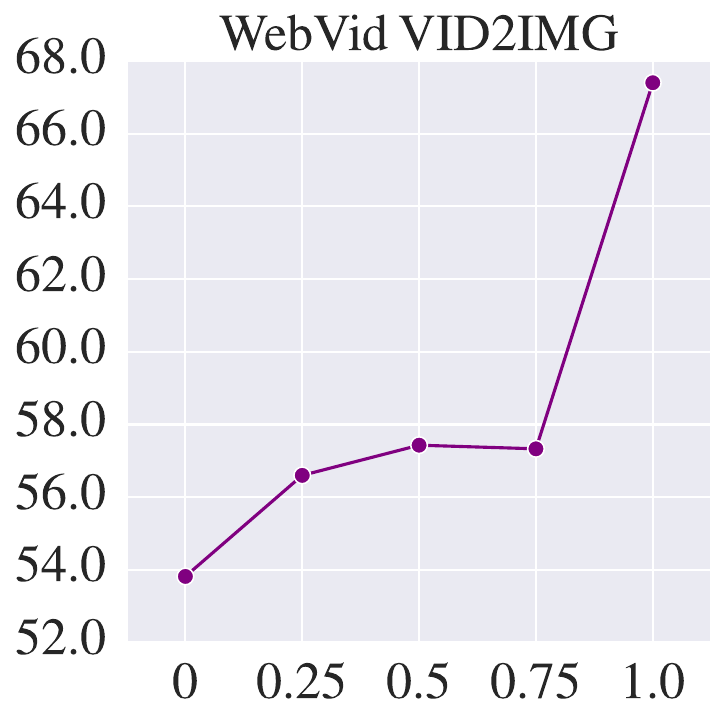}
    \vspace{7pt}
    \end{subfigure}
    \begin{subfigure}[b]{.47\textwidth}
    \includegraphics[width=\textwidth]{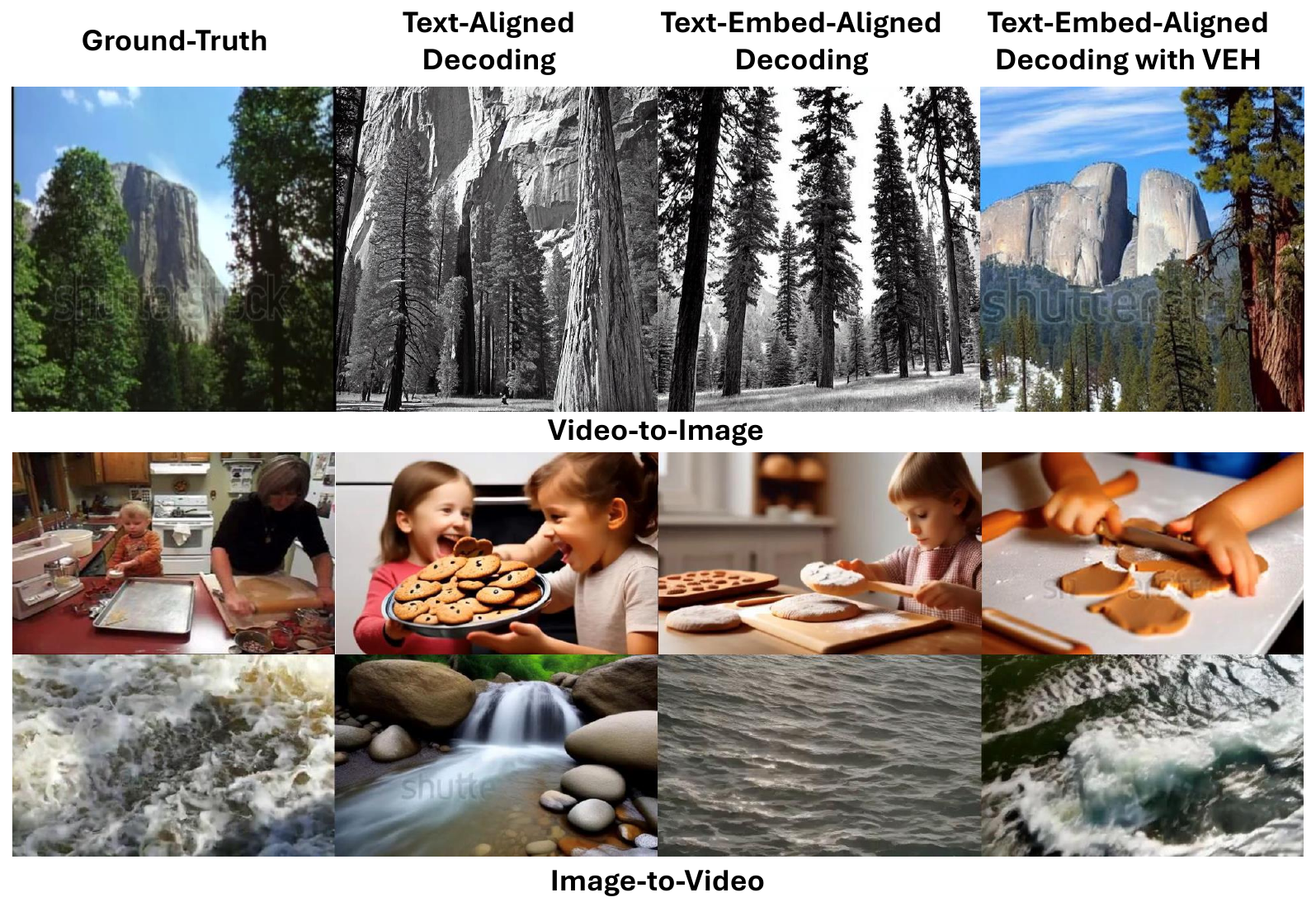}
    \end{subfigure}
    \vspace{-5pt}
    \caption{\textbf{(left, middle)} Study of using different conditioning rates in VEH (image). Higher conditioning rates brings generally better X-to-X alignment.
    \textbf{(right)} An in-depth comparison of varying design choices of X-VILA on cross-modality alignment tasks. We observe that both Text-Aligned Decoding and Text-Embed-Aligned Decoding fall short in effectively capturing semantic details from visual inputs. However, with the incorporation of our Visual Embedding Highway (VEH), we witness a substantial improvement in visual consistency.}
    \label{fig:condition_rate_an}
    % \vspace{-20pt}
\end{figure}

\subsection{Quantitative Analysis and Ablation Study}
\label{sec:main_results}
\textbf{Effectiveness of Visual Embedding Highway.} We compute the aforementioned X$^2$A scores of different models on the X-to-X alignment benchmarks of both ActivityNet Captions and WebVid datasets, and present the results in Table~\ref{tab:x2x_an} and Table~\ref{tab:x2x_webvid}. 
Specifically, we study the X$^2$A scores of Next-GPT and different versions of our X-VILA model. We investigate the performance of our model under different scenarios: \textbf{(i)} utilizing only textual alignment, \textbf{(ii)} incorporating visual alignment through the proposed visual embedding highway (VEH) on the image decoder, and \textbf{(iii)} extending VEH to both the image and video decoders.
Our findings indicate that even by utilizing textual alignment alone with our carefully curated X-to-X datasets, our model demonstrates a substantial performance advantage over Next-GPT. 
Moreover, as we progressively introduce the visual embedding highway to the image and video decoders, we observe consistent and significant improvements in visual understanding and generation tasks. 
In summary, our X-VILA demonstrates significantly stronger cross-modality understanding, reasoning, and generation ability on all types of conversation data.
These results suggest the effectiveness of our X-to-X alignment strategy and the proposed visual embedding highway design. 
Notably, both Next-GPT and X-VILA are based on the ImageBind model, making it fair to use ImageBind scores for both models.

\begin{wraptable}{r}{0.5\textwidth}
\vspace{-10pt}
  \huge
  \centering
  \resizebox{1\linewidth}{!}{
  \tablestyle{3pt}{1.2}\begin{tabular}{l ccc ccc }
  \toprule
  \multicolumn{1}{c}{\textbf{Method}}  &  \textbf{VQAv2} & \textbf{VisWiz} & \textbf{MMMU-val}    \\
  \midrule
  BLIP-2 13B~\cite{li2022blip} & 65.0 & 19.6 & -\\
  InstructBLIP 13B~\cite{Dai2023InstructBLIP} & - & 33.4 & -\\  
  Qwen-VL-Chat 7B~\cite{bai2023qwen} & 78.2 & 38.9 & 35.9 \\
  LLaVA 1.5 7B~\cite{liu2023improved} & \textbf{78.5} & 50.0 & \textbf{36.4} \\
  X-VILA 7B & 72.9 & \textbf{50.9} & 33.9 \\
  \bottomrule
  \end{tabular}
  }
   \caption{X-VILA demonstrates comparable performance to domain experts when evaluated on extra multi-modality benchmarks.
   }
\vspace{-15pt}
\label{tab:vqa_benchmarks}
\end{wraptable}

\textbf{Influence of conditioning rates.} 
We present the X$^2$A scores plotted with varying conditioning rates $\alpha$ (Equation~\ref{eq:epsilon}) in VEH (image), as depicted in Figure~\ref{fig:condition_rate_an}. Our observations indicate that an increase in $\alpha$, corresponding to more reverse steps exposed to VEH features during image sampling, leads to improved multi-modality alignment. This outcome aligns with our intuitive expectations.

\textbf{Extra multi-modality benchmarks.}
To further evaluate the multi-modality understanding capabilities of X-VILA, we perform zero-shot experiments on several multi-modality VQA benchmarks, including VQAv2~\cite{goyal2017vqav2}, VisWiz~\cite{gurari2018vizwiz}, and MMMU-val~\cite{yue2023mmmu}. 
The results in Table~\ref{tab:vqa_benchmarks} indicate that X-VILA is competitive with the leading domain-expert VLMs, while possessing the X-to-X capability.

\subsection{Qualitative Analysis and Ablation Study}
\noindent\textbf{Qualitative X-to-X alignment measurement.}
We provide a qualitative comparison to the state-of-the-art any-to-any LLMs, namely Next-GPT~\cite{wu2023next}, CoDi~\cite{tang2024codi}, and GPT-4o~\cite{openai2024chatgpt4} on visual cross-modality alignment tasks in Figure~\ref{fig:sota_compare}. 
We assess their performance by supplying an image to the models and prompting ``Please generate a video (or an image in the case of GPT-4o which cannot generate video) similar
to the semantics in the input.''
X-VILA demonstrates significant improvements in visual correspondence over previous methods, thanks to the integration of the Visual Embedding Highway (VEH) into output diffusion models.

\noindent\textbf{Emergent X-to-X ability.}
During our experiments, we observe highly promising emergent abilities displayed by X-VILA following its training on our X-to-X datasets. As depicted in Figure~\ref{fig:emergent}, we have identified two key capabilities that have surfaced: \\
(i) \textbf{Long-context cross-modality generation.} X-VILA exhibits an impressive capacity for comprehending and combining diverse concepts from multiple iterations of input. Consequently, it produces natural and coherent output, as suggested by the users.\\
(ii) \textbf{Unseen cross-modality ability.} Remarkably, X-VILA showcases the ability to perform image-to-audio and audio-to-image tasks without any explicit training on similar data. This newfound competence emerges organically through the model's exposure to our comprehensive X-to-X dataset.
These remarkable emergent abilities underscore the efficacy of our meticulously curated X-to-X dataset. Not only does it enable the model to excel in the specified data types as suggested in Section~\ref{sec:main_results}, but it also facilitates generalization across a wide range of multi-modality interactions between users and the model.

\noindent\textbf{More insights on varying design choices on decoder alignment.}
We next present our findings when aligning LLM output end to the modality-specific decoders.
We study different ways to bridge LLM output and the diffusion models: 
\textit{(i) ``Text-Aligned Decoding''}: LLM generates text description for the expected image/video/audio predictions and then feeds the text description into pre-trained image/video/audio decoders.
\textit{(ii) ``Text-Embed-Aligned Decoding''}: LLM generates modality-specific generation tokens and then we use the corresponding high-dimensional textual embeddings to control the modality-specific decoders~(as described in Section~\ref{sec:x-vila-arch}).
\textit{(iii)  ``Text-Embed-Aligned Decoding with VEH''}: Building upon method (ii), we introduce the Visual Embedding Highway~(VEH) to align the visual feature between encoders and decoders.
We conduct experiments on video-to-image and image-to-video cross-modality alignment tasks, and show the results on the right side of Figure~\ref{fig:condition_rate_an}.

The findings suggest that conveying specific details such as visual style, object appearance, and precise human actions from the input to the output is challenging for Text-Aligned Decoding. This difficulty arises due to the low-dimensional nature of pure text descriptions, which limits the amount of information they can contain. On the other hand, Text-Embed-Aligned Decoding offers a significantly greater ``bandwidth'' in the textual embedding space between the LLM and modality-specific decoders. 
Consequently, Text-Embed-Aligned Decoding is capable of generating more consistent outcomes. Nevertheless, Text-Embed-Aligned Decoding alone is still not good enough for capturing visual details, as a substantial amount of visual information is lost during the projection from encoders to the LLM. This is where our Visual Embedding Highway demonstrates its performance and aids X-VILA in attaining notably enhanced visual consistency.

\begin{figure}[t]
\includegraphics[width=1.\textwidth]{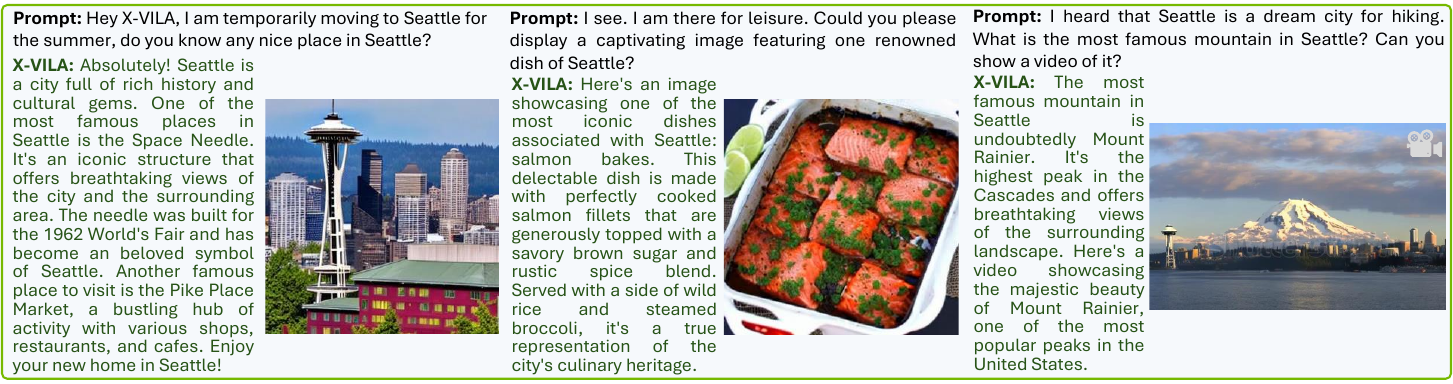}
\caption{Examples of X-VILA performing a multi-turn any-to-any modality conversation. Prompts are given left to right in a multi-round manner. 
Best viewed in color.}
\label{fig:seattle}
% \vspace{-15pt}
\end{figure}

\noindent\textbf{Conversation examples.}
To thoroughly investigate the performance of our any-to-any modality LLM, we conducted extensive testing on X-VILA examining many use cases. We present conversation examples of X-VILA across varying tasks in Figure~\ref{fig:teaser} and Figure~\ref{fig:seattle}. It can be observed that X-VILA provides users with a comprehensive set of multi-modality responses leveraging the encoders for perception, LLM for understanding and reasoning, and decoders for multi-modality content generation.
As shown in Figure~\ref{fig:supp_demo_4}, X-VILA not only exhibits its understanding of the visual input, including the scene and objects, but also predicts the actions of the person depicted in the image. This capability is a result of training on our extensive X-to-X dataset. Based on the visual input, it generates outputs visually consistent with the input, \textit{e.g.} the snow mountain and red ski suit are presented in the generation output correctly.

\section{Related Work}
\label{sec:related_work}
The era of \textbf{\textit{Large Language Models (LLM)}} arguably started with the introduction of transformers~\cite{vaswani2017attention} and a series of works that scaled them. Particularly, OpenAI introduced the Generative Pre-trained Transformer (GPT) models~\cite{radford2019language},~\cite{gpt3}, from GPT-2 (1.5B parameters) to GPT-4~\cite{GPT4} (1.76T), and showed that parameter scaling, together with more high-quality data, can generate coherent and contextually relevant text across various domains. BERT~\cite{bert} introduced a paradigm of bidirectional text processing enabling stronger context understanding and boosted question answering. T5~\cite{raffel2020exploring} converted language problem into a text-to-text format advancing translation and summarizing. Transformer-XL~\cite{dai2019transformer} demonstrated the capability of extending the context window allowing for a better understanding of longer text. The application era of LLM was kickstarted by ChatGPT~\cite{OpenAI_ChatGPT} which showcased the unprecedented ability of LLM chatbots. 

Current \textbf{\textit{Vision-Language Models (VLM)}} benefited from the development of ViT~\cite{Dosovitskiy2021AnII} that offers a unified way for vision models to communicate with other transformers from different modalities. Rapid progress has been shown in three streams~\cite{awais2023foundational}:  
\textbf{(i)} textually prompted models that accept image and text as input (CLIP~\cite{Radford2021LearningTV}, Frozen~\cite{Tsimpoukelli2021MultimodalFL}, BLIP~\cite{li2023blip}, PaLI~\cite{chen2023pali}, LLaVa~\cite{liu2023llava}, VILA~\cite{lin2023vila}, miniGPT4~\cite{zhu2023minigpt}); 
\textbf{(ii)} visually prompted models (CLIPSeg~\cite{Luddecke2021ImageSU}, SAM~\cite{kirillov2023segment}); and \textbf{(iii)} multi-modal input-output models (Painter~\cite{Wang2022ImagesSI}, ImageBind~\cite{Girdhar2023ImageBindOE}, Palm-E~\cite{Driess2023PaLMEAE}, Video ChatGPT~\cite{maaz2023video}, RegionGPT~\cite{regiogpt}, mPLUG-owl~\cite{ye2023mplug}, PandaGPT~\cite{su2023pandagpt}, CoDi~\cite{tang2024codi}, NextGPT~\cite{wu2023next}, Unified-IO~\cite{lu2022unified,lu2023uio2}). Among the first, Frozen~\cite{Tsimpoukelli2021MultimodalFL} demonstrated that VLM can be constructed by linear projection of ViT features into LLM and only tuning ViT on image-text captioning data. They are the first that discover the few-shot capabilities of VLM without instruction. Flamingo~\cite{alayrac2022flamingo} used cross-attention for vision language binding, and for a first time demonstrated surpassing state-of-the-art finetuned models for multiple tasks. PALI~\cite{chen2023pali} created a universal model that can do vision and language tasks separately, they scaled ViT to 4B and demonstrated the importance of adding language-only data to the pretraining stage. Overall, VLM follows the pipeline of taking a pretrained LLM; adding a pretrained vision encoder; learning feature alignment at scale via a projector or cross-attention; followed by instruct-tuning (InstructBLIP~\cite{Dai2023InstructBLIP}, FLAN~\cite{wei2021finetuned}). 
In close relation to our research, Next-GPT introduces an LLM that possesses the capability to comprehend multi-modality inputs and generate corresponding multi-modality outputs through textual alignment, yet it cannot effectively handle visual details present in the input.

\section{Conclusion}
This paper presents X-VILA, an any-to-any modality LLM that is able to understand, infer, and generate multi-modality contents. This ability is achieved through any-to-any modality alignment, for which we curate a dataset for any-to-any modality instruction tuning. 
We further identify a significant drawback in the previous textual alignment method that leads to the loss of crucial visual details. Accordingly, we propose an innovative visual alignment mechanism that incorporates a visual feature highway module. This solution helps preserve essential visual details from the input.
The experimental results, both quantitative and qualitative, indicate the effectiveness of our data and methodology. 
X-VILA's performance can be further enhanced across various VLM benchmarks.

\bibliographystyle{unsrt}
\bibliography{main}

\clearpage
\appendix

\begin{appendix}

\newcounter{mysubtable}
\newcommand\modcounter{%
  \refstepcounter{mysubtable}%
  \renewcommand{\thetable}{\Alph{mysubtable}}%
}
\newcounter{mysubfig}
\newcommand\modfigcounter{%
  \refstepcounter{mysubfig}%
  \renewcommand{\thefigure}{\Alph{mysubfig}}%
}

\title{X-VILA: Cross-Modality Alignment for\\Large Language Model \\ 
\rule[0.25\baselineskip]{0.5\textwidth}{1pt}\\
Supplemental Material}

\clearpage

\section{X-VILA Training}
\label{sec:x-vila-training}
The training process of X-VILA is divided into three phases, namely (i) encoder-LLM-Decoder alignment training, (ii) interleaved data pre-training, and (iii) X-to-X cross-modality instruction fine-tuning. 

\subsection{Encoder-LLM-decoder alignment training phase.}

As the first step, we align the output of modality-specific encoders and the input of modality-specific decoders to the textual embedding space of LLM, as detailed in~\cite{wu2023next}. 
To achieve this goal, we only train the input projection layers, output projection layers, and the vocabulary embedding layer of LLM, while keeping all other parameters frozen.
We use corpus with ``X''-text pairs to train the model, where ``X'' is one of the video, image, or audio modalities.

For this stage, we design two primary tasks to train the projection layers: X-to-text generation and text-to-X generation.
\textbf{(a)} X-to-text generation includes video, image, and audio captioning tasks. The model is supervised to generate text based on the multi-modality inputs. During this process,
 the input projection layers are trained to align the output embedding of modality-specific encoders and the textual embedding space of pre-trained LLM. 
\textbf{(b)} Text-to-X generation aims at aligning the output textual embedding space of LLM and the input end of modality-specific decoders. We use video, image, and audio generation tasks to train the model, where only the output projection layers are optimized.
As previously mentioned, the training objective here is pure textual alignment: minimizing the feature distance between the textual controller embedding $\mathbf{E}^{\text{text}}_m$ generated by the output projection layers and the embedding generated by the original pre-trained text encoder of diffusion model.  This training strategy ensures that $\mathbf{E}^{\text{text}}_m$ shares a distribution similar to that of the pre-trained text encoder in the diffusion model. After training, $\mathbf{E}^{\text{text}}_m$ replaces the diffusion text encoder feature to control the U-Nets of the modality-specific decoders via cross-attention. 

\subsection{Interleaved data pre-training phase.}
Interleaved data training has been proven to be an effective strategy for vision-language models in alleviating the catastrophic forgetting issue after training on only visual-text pairs, and obtaining long-context understanding ability~\cite{lin2023vila,openflamingo}.
Therefore, we introduce a dedicated phase for pre-training X-VILA using a multi-modality interleaved corpus. In addition to interleaved image-text pairs as in MMC4~\cite{zhu2023multimodal}, we further construct a new dataset from ActivityNet Captions~\cite{krishna2017denseANC}. The main idea is to exploit the nature of video that contains sequential flow of text (\eg, captions), audio, short video, and image. This enables us to put the images/videos and texts in an interleaved manner, and use the corpus to pre-train X-VILA. 

Specifically, we construct interleaved multi-modality data sequences from each target video clip as:

\begin{align}
\scriptsize \nonumber
    \underbrace{\texttt{\{<img. 1>, <aud. 1>, <vid. 1>, <txt 1>\}}}_{\textrm{sampled from video chunk } 1}, ..., \underbrace{\texttt{\{<img. n>, <aud. n>, <vid. n>, <txt n>\}}}_{\textrm{sampled from video chunk } n},
\end{align}
where the video chunks are sampled from an entire video clip that offers natural sources of interleaved cross-modality data structure. Once constructed, the modalities are sampled during training to align varying targets for gradient computation and network projector alignment. In this work, we observe the even sampling method and $n=3$ are sufficient for the task, namely constructing cross-modality tasks for the beginning, middle stage, and ending of video clips. 
During this stage, we jointly train the input and output projection layers, and use LoRA~\cite{hu2021lora} on LLM for fine-tuning.

\subsection{X-to-X cross-modality instruction tuning phase.}
\begin{figure}[t]
    \centering
     \includegraphics[width=\textwidth]
    {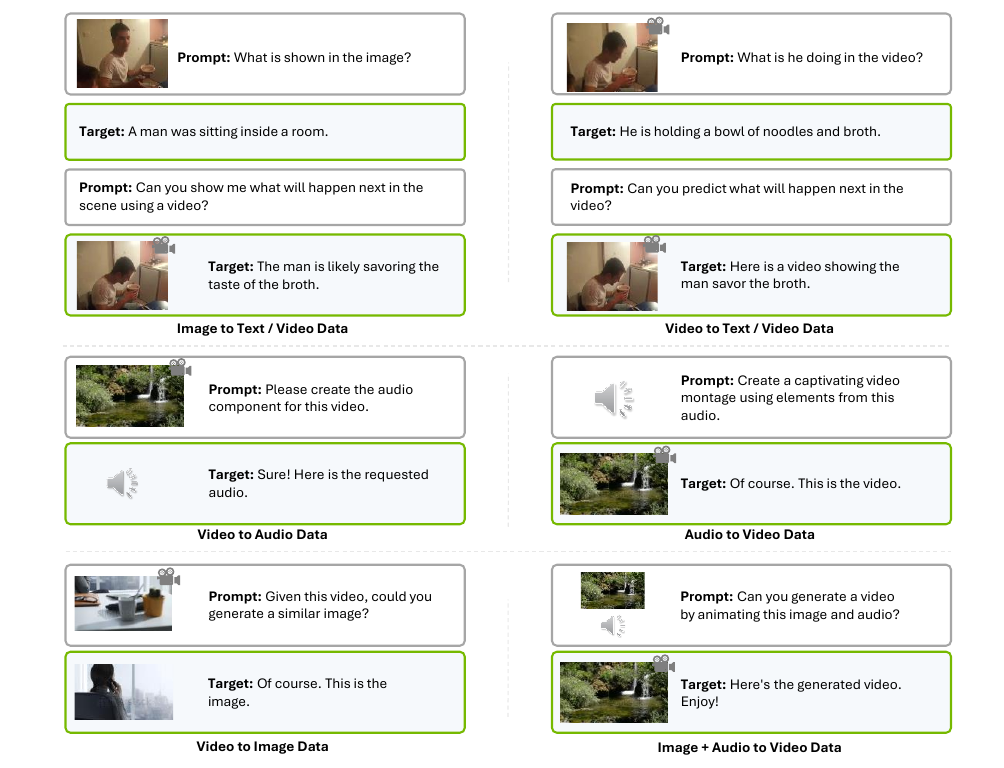}
    \caption{Examples of different types of conversations in our X-to-X dataset. They cover six types of cross-modality understanding and generation tasks.
        }
    \label{fig:x2xdataset}
\end{figure}

After the previous two phases, we have textually aligned different components of X-VILA in a unified framework. However, the model is still not ready for understanding and generating multi-modality content in a proper manner. 
To achieve this goal, we curate a comprehensive ``X-to-X dataset'' for cross-modality generation instruction tuning.
As video captioning datasets are inherently multi-modal and provide abundant corpus in video, audio, image, and text forms, we build our X-to-X dataset based on two video captioning datasets: Webvid~\cite{Bain21webvid} and ActivityNet Captions~\cite{krishna2017denseANC}. 
Our X-to-X dataset features six different types of cross-modality generative conversations, namely \textbf{video-to-image, video-to-video, image-to-video, video-to-audio, audio-to-video, and image+audio-to-video}.
We show examples of different types of conversations in Figure~\ref{fig:x2xdataset}.
Each conversation contains one or more rounds of cross-modality conversation.
More details about the X-to-X dataset are described in the experiment section.

We further divide the X-to-X cross-modality instruction tuning phase into two distinct steps, each based on different alignment methods: textual alignment and visual alignment.

\noindent \textbf{(a)} 
To achieve textual alignment, we first project the multi-modality inputs into the textual embedding space of LLM. Then, LLM generates textual embeddings that are subsequently converted into the corresponding modality's content.
We follow a process similar to phases (i) and (ii). Firstly, for image, video, or audio outputs, we generate embeddings using the text encoders of corresponding diffusion models. We then optimize the distance between these embeddings and the $\mathbf{E}^{\text{text}}_m$ generated by our model. During this step, we keep all the decoder weights frozen and train the input projection layers, output projection layers, and vocabulary embedding layer as well as LoRA parameters of LLM.
For training data, we blend our X-to-X dataset with common SFT datasets used by other VLM models~\cite{liu2023llava,wu2023next} (more details in the experiment section).

\noindent \textbf{(b)} 
As mentioned earlier, relying solely on textual alignment is inherently insufficient to retain the visual details of the input when generating visual outputs.
To address such an issue, we design a novel visual alignment method. We propose a visual embedding highway~(VEH) module as introduced in Section~\ref{sec:x-vila-arch}, which is utilized for the image and video decoders when there is a visual modality in the input. During training, we update the parameters of the visual decoders and the visual controller module. Meanwhile, we keep all other network parameters fixed, including the input and output projection layers and LLM. In this way, the model's ability to conduct tasks in other modalities is not influenced by the visual alignment process.

\section{More Qualitative Results}

\subsection{Examples of our X-to-X dataset.}
To provide an intuitive understanding of the six types of conversations in our curated X-to-X dataset, we visualize the conversation samples of the dataset in Figure~\ref{fig:x2xdataset}.
The design of the dataset focuses on building any-to-any modality connection through various conversation templates.

\subsection{Visual comparison with CoDi on cross-modality alignment.}
\begin{figure}[h]
    \centering
    \includegraphics[width=.98\textwidth]{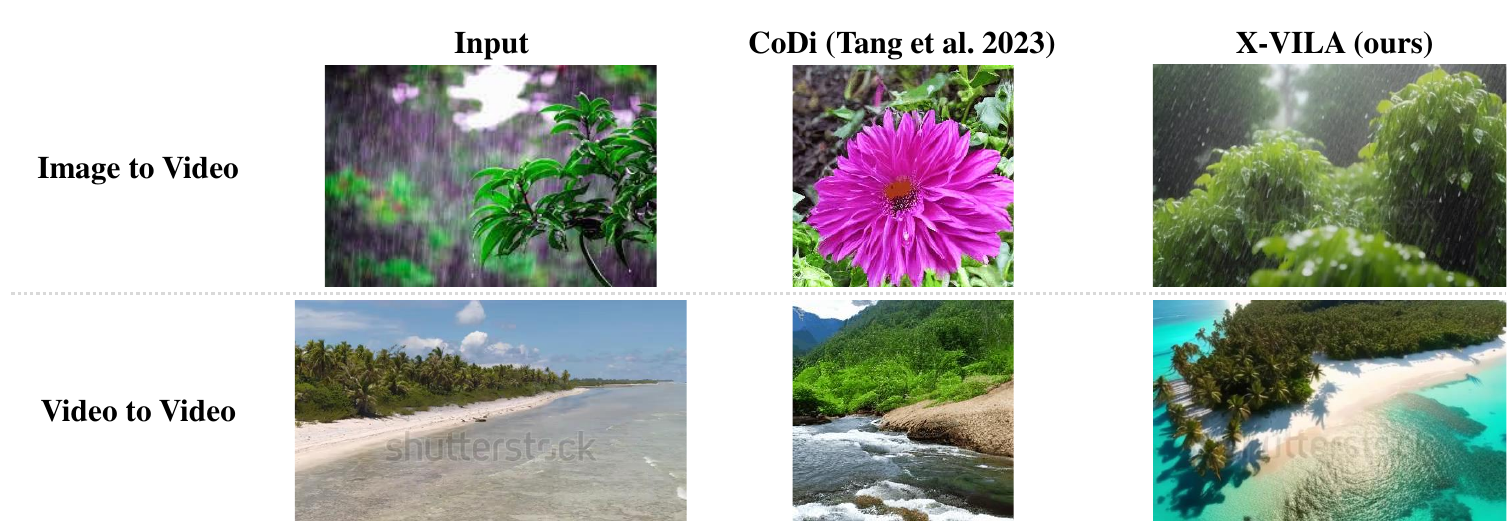}
    \caption{Visual comparison to the recent work CoDi~\cite{tang2024codi} on cross-modality alignment. X-VILA demonstrates largely improved generation quality and cross-modality consistency.
   }
    \label{fig:sota_compare_codi}
\end{figure}

To further examine the visual alignment advantage of X-VILA, we compare it with the state-of-the-art any-to-any model CoDi~\cite{tang2024codi} in Figure~\ref{fig:sota_compare_codi}. We observe that CoDi fails to capture the real semantics and details of the input. Notably, CoDi is unable to perform X-to-X chatting, unlike X-VILA, which is specifically designed for omni-modality chatting while being able to produce superior visually aligned generation results.

\subsection{Human-model interaction demonstration.}
\begin{figure}[t]
    \centering
    \vspace{-10pt}
    \includegraphics[width=\textwidth]{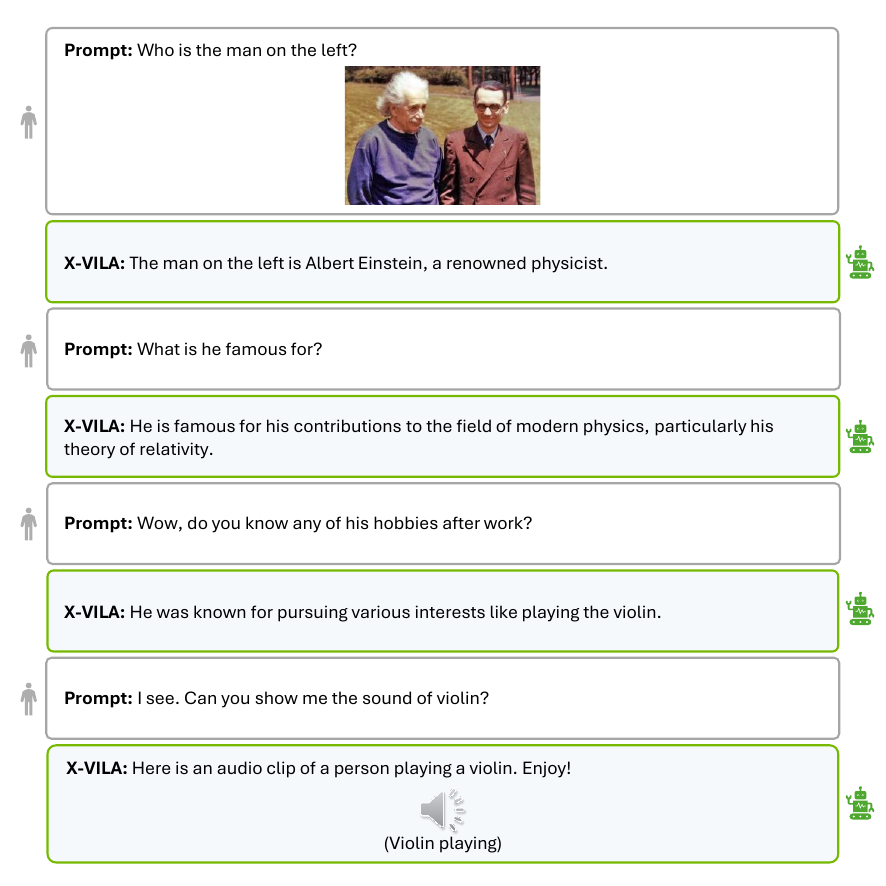}
    \vspace{-15pt}
    \caption{Example of X-VILA performing cross-modality chat. X-VILA shows strong multi-understanding and generation ability.
   }
\label{fig:supp_demo_1}
\vspace{-10pt}
\end{figure}

\begin{figure}[t]
    \centering
    \vspace{-10pt}
    \includegraphics[width=\textwidth]{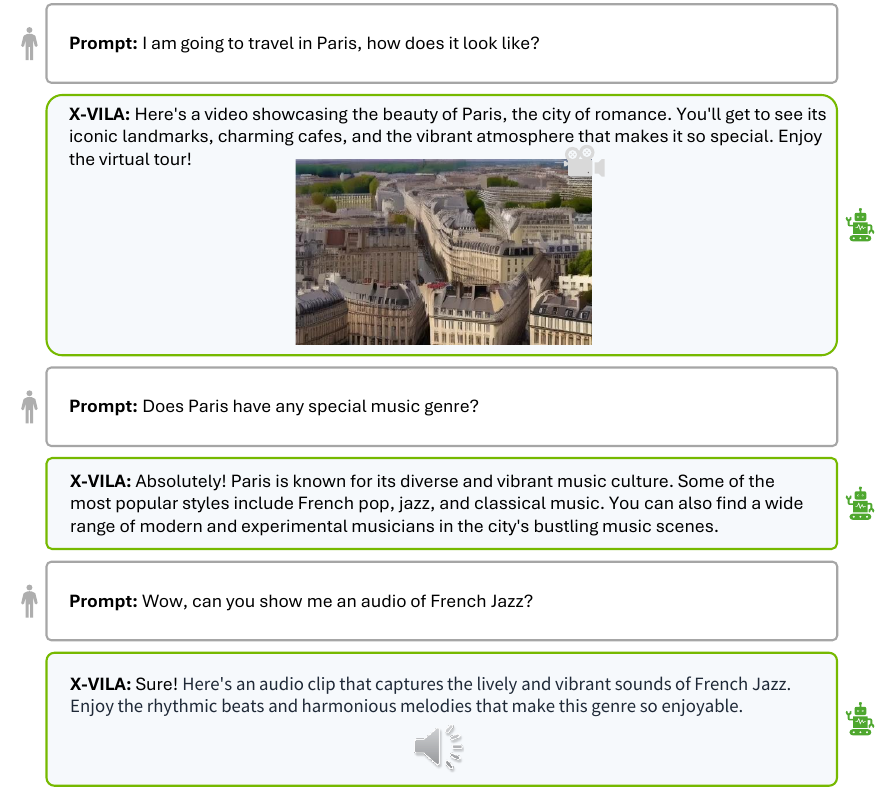}
    \vspace{-15pt}
    \caption{Example of X-VILA performing cross-modality chat. X-VILA shows strong multi-understanding and generation ability.
   }
\label{fig:supp_demo_2}
\vspace{-10pt}
\end{figure}

\begin{figure}[ht]
    \centering
    \includegraphics[width=.98\textwidth]{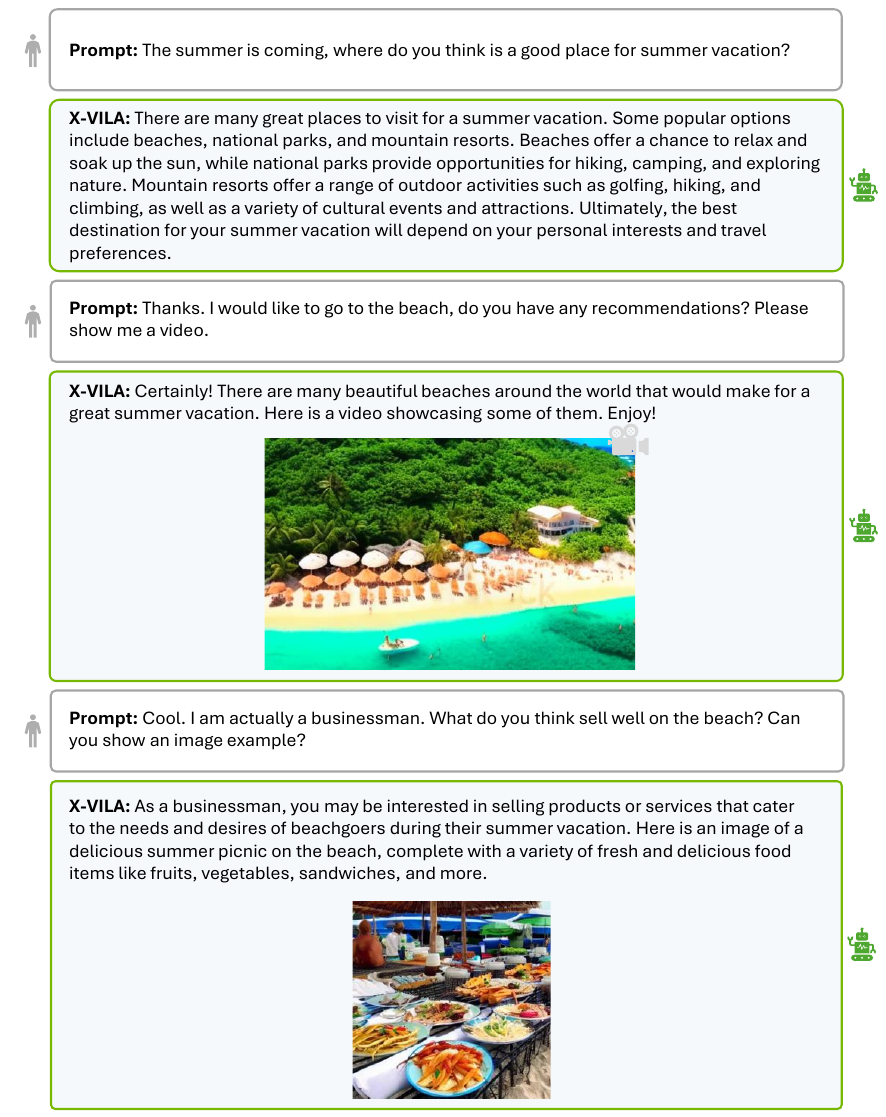}
    \caption{More examples of X-VILA performing cross-modality chatting.
   }
\label{fig:supp_demo_3}
\vspace{-8mm}
\end{figure}

\begin{figure}[ht]
    \centering
    \vspace{-6mm}
    \includegraphics[width=.98\textwidth]{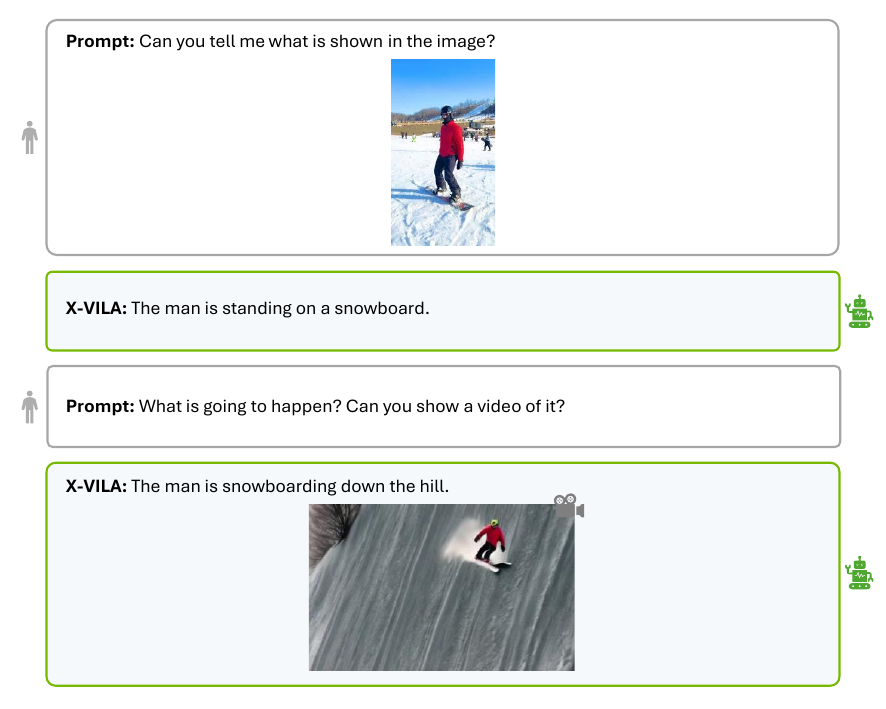}
    \vspace{-5pt}
    \caption{Conversation examples of X-VILA performing cross-modality conversation. X-VILA demonstrates a remarkable capability to comprehend the visual input and perform reasoning based on it. Our proposed visual alignment technique plays a crucial role in establishing visual consistency between the input and output.
   }
\label{fig:supp_demo_4}
\end{figure}

To conduct a comprehensive assessment of our any-to-any modality LLM's performance, we undertake more testing on X-VILA, meticulously examining different use cases. We present a collection of human-model conversation examples in Figure\ref{fig:supp_demo_1}, \ref{fig:supp_demo_2}, \ref{fig:supp_demo_3} and \ref{fig:supp_demo_4}, showcasing the versatility of X-VILA across diverse tasks.
These results demonstrate the effectiveness of X-VILA in addressing the needs of users by offering comprehensive and generative multi-modality capabilities.

\section{More Implementation Details}
As introduced in Section~\ref{sec:x-vila-training}, X-VILA training is separated into three phases.
(i) In the initial phase, referred to as encoder-LLM-decoder alignment training, the model undergoes 20,000 iterations using an Adam optimizer. The base learning rate is set to $4\times 10^{-4}$, and a learning rate warm-up strategy is employed. The batch size for this phase is set to 200.
(ii) During the second phase, known as interleaved data pre-training, a batch size of 192 is utilized. The base learning rate is set to $1\times 10^{-4}$, and the training is conducted for 10,000 iterations.
(iii) The final phase, called cross-modality instruction tuning, involves separate training for textual and visual alignment. For textual alignment, a batch size of 192 is maintained, and the model is trained for 30,000 iterations using a base learning rate of $1\times 10^{-4}$.
Regarding visual alignment, both the Visual Embedding Highway (VEH) and modality-specific decoders are trained for 20,000 iterations. The batch size for this phase is set to 64, and the learning rate is adjusted to $1\times 10^{-6}$.
For data amount, 
Our training pipeline is highly efficient compared to many previous vision-language models~\cite{alayrac2022flamingo,bai2023qwen,chen2023pali}.
We utilize a total of 4 NVIDIA A100 80GB server nodes in the training process.

\clearpage

\end{appendix}

\end{document}